# Using the Full-text Content of Academic Articles to Identify and Evaluate Algorithm Entities in the Domain of Natural Language Processing

Yuzhuo Wang, Chengzhi Zhang *

Department of Information Management, Nanjing University of Science and Technology, Nanjing, 210094, China

**Abstract:** In the era of big data, the advancement, improvement, and application of algorithms in academic research have played an important role in promoting the development of different disciplines. Academic papers in various disciplines, especially computer science, contain a large number of algorithms. Identifying the algorithms from the full-text content of papers can determine popular or classical algorithms in a specific field and help scholars gain a comprehensive understanding of the algorithms and even the field. To this end, this article takes the field of natural language processing (NLP) as an example and identifies algorithms from academic papers in the field. A dictionary of algorithms is constructed by manually annotating the contents of papers, and sentences containing algorithms in the dictionary are extracted through dictionary-based matching. The number of articles mentioning an algorithm is used as an indicator to analyze the influence of that algorithm. Our results reveal the algorithm with the highest influence in NLP papers and show that classification algorithms represent the largest proportion among the high-impact algorithms. In addition, the evolution of the influence of algorithms reflects the changes in research tasks and topics in the field, and the changes in the influence of different algorithms show different trends. As a preliminary exploration, this paper conducts an analysis of the impact of algorithms mentioned in the academic text, and the results can be used as training data for the automatic extraction of large-scale algorithms in the future. The methodology in this paper is domain-independent and can be applied to other domains.

## 1. Introduction

The speed of social development is accelerating, and new technologies are born every day. Societal developments provide people with new opportunities and conveniences. However, human beings still face new challenges and problems. At the end of 2019, a novel coronavirus (SARS-CoV-2) was detected in China. The virus spreads very quickly, causing substantial losses to the entire society. Issues such as how to find a cure for the virus, how to find the source of the virus, how to develop a vaccine, and how to distribute materials during the epidemic require experts and scholars to find new or more suitable methods in their research field.

Among different categories of methods, algorithms are bound to have an important role; algorithms are ubiquitous and offer precise methodologies to solve problems (Carman, 2013). Informally, an algorithm is any well-defined computational procedure that takes a set of values as input and produces some value as output (Cormen et al., 2009), which is needed in scientific research. Especially in the era of big data, data-driven research requires algorithms to extract, process, and analyze massive amounts of data. Therefore, algorithms have become research objects, as well as useful technologies, of scholars in different fields. In the "Venice Time Machine" project in the field of digital humanities, researchers used machine learning algorithms to reveal Venice's history in a dynamic digital form to reproduce the glorious style of the Republic of the ancient city (Abbott, 2017). In the field of computer science, scientists have used machine learning algorithms to combat the novel coronavirus, including the use of algorithms to detect infections, differentiate COVID-19 from the common flu and to predict the epidemic situation (Dave, 2020).

Academic papers in many disciplines, especially in the computer science domain, propose, improve, and use various algorithms (Tuarob & Tucker, 2015). However, not everyone is an algorithm expert. For many researchers, especially beginners in a field, gaining a thorough understanding of the algorithms and finding one that is suitable for their own research are urgent problems. Scholars usually find suitable algorithms through two methods. One method is direct consultation with more experienced scholars, but this method depends on the advisers' knowledge and does not guarantee the comprehensiveness of the algorithm suggestions. Another method is reading academic literature and finding algorithms from the research of other people, which provides scholars with more algorithms. Academic papers are a perfect source of algorithms; however, information overload cannot be ignored. The research has pointed out that the number of academic literature entries generated worldwide has reached the level of millions, and it continues to increase at a rate of approximately 3% each year (Bornmann & Mutz, 2015). If scientists only search for algorithms by reading articles, it will be a time-consuming and labor-intensive challenge. If the algorithms mentioned in papers, namely, any algorithm appearing in the papers, including the algorithm proposed, used, improved, described or simply mentioned by the author, can be identified and evaluated, it can save time for scholars and provide a solid foundation for them to sort out the algorithms of specific disciplines or research topics.

To this end, this article aims to collect algorithms in research papers in a domain and further explore the influence of algorithms. Tuarob et al. (2020) defined a standard algorithm in academic papers as one that is well known by people in a field and is usually recognized by its name, including Dijkstra's shortest-path algorithm, the Bellman-Ford algorithm, the Quicksort algorithm, etc. On this basis, we use our experience, authors' descriptions and other external knowledge to annotate the named algorithms in articles. In addition, we posit that, when an algorithm appears in an article, it has an influence on the article. Therefore, we evaluate the influence of an algorithm based on the number of papers that mention the algorithm in the full-text content. Mention count has proven to be a suitable indicator to measure the influence of entities in academic papers (Howison & Bullard, 2016; Ma & Zhang, 2017; Pan et al., 2015). Therefore, we take the field of natural language processing as an example and explore three research questions:

*RQ1*. What are the high-influence algorithms in natural language processing?
*RQ2*. What are the differences in the influential algorithms in different years?
*RQ3*. How does the influence of the algorithm change over time?

To be more specific, we attempt to research on the influence of different algorithms in NLP domain through the first question. For question 2, we combine the influence and time to understand the changes of high-impact algorithms in different years, and try to analyze the development of NLP field from the perspective of changes in algorithm. The third question is the refinement of the second question. We will explore evolution of specific independent algorithms and give the pattern of changes in influence.

The reason for choosing the field of natural language processing is that papers in the computer science field are more likely to propose and use algorithms than papers in other fields. Furthermore, computer science is a rapidly developing discipline, and there are various algorithms that emerge in the discipline, which ensures that we can collect enough algorithms to carry out our research. It should be noted that, in the traditional named-entity recognition task, named entities refer to nouns or noun phrases representing various entities (Petasis et al., 2000). Therefore, in this paper, the annotated algorithm refers to the noun or noun phrase representing the algorithm with a specific name, for example, the *support vector machine*, rather than the concept described by the author, for example, a novel classification algorithm.

## 2. Related Works

The algorithm entity is a kind of knowledge entity. To be specific, the existing research classifies knowledge entities into various types, including research methods, theories, software, algorithms, datasets and so on. Scholars use the full text of academic papers to identify entities by manual annotation or machine learning methods and then analyze the influence of knowledge entities based on various indicators.

### 2.1 Evaluating knowledge entities based on frequency

At present, most papers analyze the influence of knowledge entities based on bibliometric indicators, which are usually the frequency of mentions, citations and uses of entities in academic papers (Belter, 2014). Jarvelin and Vakkari (1990) pioneered the use of academic papers to collect research methods and establish a methodological framework. Blake (1994) turned his attention from journals to dissertations. By reading the contents of abstracts, he identified research methods often used in dissertations. Pan et al. (2015) adopted the method of bootstrapped learning to automatically extract software entities from academic articles and evaluated the software according to the frequency of citation and use. He et al. (2019) selected 14 science mapping tools and analyzed their influence by exploring the number of citations of articles that used these tools. On the basis of frequency, scholars also gave some other indicators to analyze the influence of knowledge entities from different aspects. Chu and Ke (2017) annotated the research methods in academic papers of LIS (library and information science) and combined the frequency of mention with the time of publication to analyze the changes in the influence of research methods. Pan et al. (2016) studied the influence of knowledge entities in different disciplines, and the results show that software is more widely used in agriculture, health sciences, and biology than mathematics, information sciences, and social sciences. Zhao et al. (2018) manually annotated the datasets in PLoS One papers and analyzed the collection, storage, availability and other features of data sets in various disciplines based on frequency.

In addition to bibliometric indicators, some altmetric concepts are also used to evaluate the influence of entities. These indicators can be the frequency of votes, downloads, and visits of entities. In 2006, the organizers of ICDM (The IEEE International Conference on Data Mining) used the votes of experts to evaluate the influence of algorithms (Wu et al., 2008). Stack Overflow also evaluates the influence of IT technology and databases based on the votes of practitioners in the IT domain [1]. TIOBE considers the number of programmer votes, the number of courses and the number of vendors to calculate the popularity and influence of different programming languages[2]. Thelwall and Kousha (2016) posited that the download volume of open-source software can be used as an indicator of software value. Based on the concept, Priem and Piwowar initiated the open-source project Depsy[3], which is dedicated to analyzing the influence of various software in the open-source community. The indicators include the frequency of downloads, citations in academic papers and so on. Subsequently, Zhao and Wei (2017) used Depsy to obtain the number of downloads and citations of some software in the Python community and analyzed the academic influence. Generally, different types of frequencies have their own advantages. How to integrate them to analyze the influence of knowledge entities is worth exploring in the future.

**2.2 Evaluating knowledge entities based on text content**
In addition to using frequency to measure influence, some works have also utilized text content to deeply explore the role, function and relationship of knowledge entities. Li et al. (2017) took 19,478 papers that mentioned WoS (web of science) as the research object and analyzed the abstract content through Stanford CoreNLP. According to the verbs and language patterns in the context related to WoS, the results showed that the most important reason researchers mention WoS is that it is used as a source of data. The analysis of the relationship includes both the relationship between entities and the relationship between entities and articles. Li and Yan (2018) proposed a software recognition algorithm based on the software name dictionary. By extracting sentences that mentioned the R software package in papers published by PLOS, they analyzed the co-mention network of the software package and noted that packages with similar disciplines and functions were more likely to be mentioned at the same time. Similarly, Zhang et al. (2019) used the full text of academic papers published by PLOS One to cluster 260 kinds of software mentioned in the article. The results also showed that software with similar functions would be clustered together. Yang et al. (2018) analyzed the relationship between article and software, and the results indicated that articles published by a journal with higher quality tended to use newer software and that international articles used new software earlier than Chinese articles.

Furthermore, some scholars have used the content of academic papers to identify the pattern of citation and use of knowledge entities. Yoon et al. (2019) investigated how the Health Information National Trends Survey (HINTS) data were cited in academic literature. The results indicated that more than half of the articles cited HINTS-related documents rather than the data itself. Costa et al. (2018) took statistical analysis tools as the example and researched the sustainability of statistical analysis tools. The results demonstrated that many tools had short life cycles. Similar to datasets, software has a diverse citation pattern in academic articles. Pan et al. (2019) studied the software that was actually used in academic papers rather than just mentioned in the LIS (library and information science) discipline, and they pointed out that the dependence of papers on software was increasing gradually and that the citation pattern of software was various and irregular. The work of Li et al. (2017) indicated that the reason for the informal citation of software is that the citation standards of software were diversified and that the authors have not followed the specifications.

---

[1] https://insights.stackoverflow.com/survey/2020/?utm_source=social-share&utm_medium =social&utm _campaign =dev-survey-2020
[2] https://blog.okfn.org/category/open-data-index/
[3] http://depsy.org/

**2.3 Application of extracted knowledge entities**

Tuarob et al. (2013) carried out a series of applications based on knowledge entities. They first used rule-based method to identify algorithms in academic papers, and built an algorithm search system (Bhatia et al., 2011). At the same time, with the help of the full text, the co-citation network (Tuarob et al., 2012), function (Tuarob et al., 2013) and efficiency (Safder et al., 2017) of algorithms were further investigated, which can be used to optimize the search system. With the development of technology, they combined the rule-based methods and machine-learning methods, and by identifying the pseudocode and process description in the full text of academic literature, algorithms mentioned in the article were extracted. After that, they built the novel algorithm search system, namely algorithmseer (Tuarob et al., 2016), and gave the query results by calculating the similarity between the algorithm description and the search words. Citation function of algorithms was explored to improve the performance of the system (Tuarob et al., 2020). In addition to the search system, Zha et al (2019) utilized the deep learning method to extract algorithms from the tables in academic papers, and then constructed an algorithm roadmap to describe evolutionary relation between different algorithms. However, these studies have few limitations. They only target the algorithms with restrictions. For example, the algorithms proposed by the authors, the algorithms described in detail, or algorithms mentioned in tables. The algorithms without detailed description or appearing in other parts of the article were ignored.

In general, the existing work on extraction and evaluation of knowledge entities has received widespread attention. For the assessment of the influence of knowledge entities, frequency is still the main indicator. In addition to basic frequency-based impact assessments, some work has explored features such as functions, relationships, uses and citation patterns of knowledge entities. Among different knowledge entities, software gets the most attention because software has a clearer definition and is easier to identify. However, there are few research studies on algorithms. Even if studies have identified entities and built a retrieval system and roadmap, they only focused on the algorithms proposed in the article or the algorithms described in detail, and did not evaluate the influence of algorithms. To this end, this article attempts to identify more algorithm entities from papers and construct an algorithm dictionary by manual annotation. We intend to conduct preliminary statistical analysis of the algorithms mentioned in academic papers in specific fields and carry out more in-depth exploration based on the data in the future.

# 3. Methodology

As shown in Figure 1, using natural language processing (NLP) as a case, we identified and evaluated algorithms in the full-text content of academic articles in the NLP domain. For the full text of academic papers, we identified algorithms and compiled the dictionary manually, and then we extracted the algorithm sentences from articles through dictionary-matching and evaluated the influence of different algorithms. Different from the bootstrapping method used in other work, we used the method of manual annotation to identify the algorithm entities, which could ensure the accuracy of the recognition results.

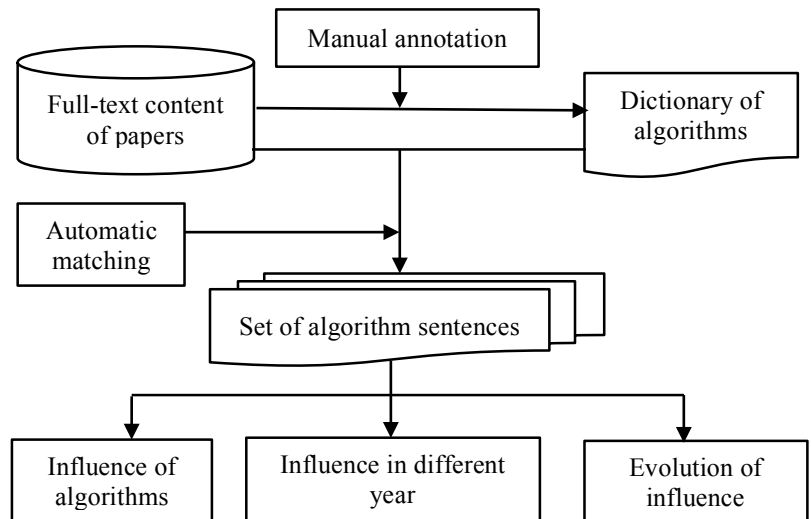

Figure 1. Framework of our work

### 3.1 Data collection

We selected conference proceedings of the ACL (Annual Meeting of Association for Computational Linguistics) as the dataset. In the computer science field, outstanding research results are often published in conferences rather than journals, and therefore, conference papers are more suitable resources for exploring the algorithms in computer science. ACL is the top conference in the field of NLP, and it is believed that research methods and results in ACL conference papers are more representative. To this end, papers published in the ACL conference could be an excellent material for studying algorithms in NLP. We downloaded all of the ACL conference papers published between 1979 and 2015 from the ACL anthology reference corpus (https://acl-arc.comp.nus.edu.sg/), and 4,641 papers were available in both PDF and XML formats.

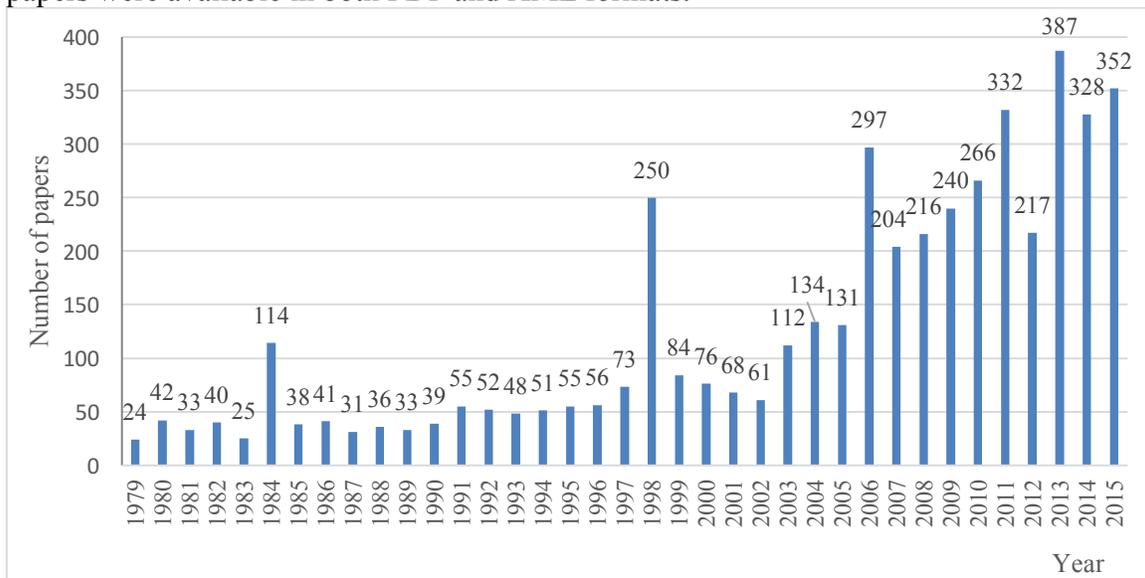

Figure 2. The number of papers accepted by ACL conference each year

Figure 2 shows the number of papers published each year in the ACL conference. It is clear that, before 2003, the number of papers accepted by the conference was less than 100 per year. The year 1998 was a special year, with two volumes of conference proceedings published. Since 2006, the number of ACL conference papers has increased significantly.

**3.2 Manual annotation of algorithm entities in academic papers**

For the papers in PDF format, we randomly selected a sample of 100 papers from the entire dataset. Two annotators, a Ph.D. student and a master student, were invited to annotate the algorithm entities mentioned in the 100 papers. Both of the annotators were familiar with natural language processing. The labeling process included the following steps:

(i) The annotators read the title, abstract, introduction, method and result of the article and labeled algorithms mentioned in the content of these sections.

(ii) The annotators reviewed figures and tables in the article since the results of the used algorithms are usually presented in tables and figures.

(iii) The annotators used the word "algorithm" as a search term to perform automatic retrieval in the full text to find algorithms in other sections.

(iv) The annotators quickly browsed the full text to determine whether there were any missing algorithms.

In the process of tagging, the annotators used the following steps to determine whether a noun or noun phrase should be labeled as an algorithm:

(i) For a well-known algorithm, or if all of the authors of the ACL papers called the method an "algorithm", it was directly labeled as an "algorithm", for example, support vector machines.

(ii) For a noun that the author did not define in the paper but was suspected to be an algorithm, for example, the Gibbs sampling method, the annotators searched for it in external knowledge bases (Wikipedia, Google Scholar, other academic papers or monographs) and determined whether it was an algorithm according to the introduction.

(iii) For a noun for which the author did not provide a unified definition, further judgment was required. Taking the Hidden Markov Model as an example, some authors called it a "model," and some authors called it an "algorithm." After accessing other information and consulting experts, the annotators determined that in essence, it is still an "algorithm."

We combined the results independently labeled by the two annotators and then compiled an algorithm dictionary. After that, we used the method of dictionary-matching to match algorithms in the dictionary with the full-text content of the 100 articles. Thus, we could find algorithms that were ignored by the annotators in some articles but found in other articles. In the end, we knew which algorithms from the dictionary appeared in each article, and we regarded the result as the gold standard and compared it with the two annotators' original annotations. Taking each article as a unit, we identified the algorithms that each annotator ignored in each article and calculated the missing rate. The missing rates of the two annotators were 13% (Ph.D. student) and 14% (Master student), and both of the coders missed 11% of algorithms in the gold standard. To measure the interrater reliability (IRR) between the two labelers, we employed Cohen's kappa coefficient (Cohen, 1960) and achieved an IRR of 0.78. The result indicates the sufficient reliability of one labeler annotating all of the papers. Therefore, the Ph.D. student annotated all of the remaining papers.

The results were stored in tables. As shown in Table 1, we recorded the ID (the unique identifier consisting of numbers and letters for each paper in the ACL anthology reference corpus), title of the paper and all of the algorithms appearing in the body text.

Table 1. Example of algorithms extracted from articles

| ID | Title | Alg. 1 | Alg. 2 | Alg. 3 | Alg. 4 | Alg. 5 | Alg. 6 |
|---|---|---|---|---|---|---|---|
| P01-1049 | Building Semantic Perceptron Net for Topic Spotting | Naïve Bayes | KNN | SVM | NNet | LSF | BP algorithm |

### 3.3 Algorithm dictionary compilation

Since authors' writing styles are different, algorithms may be mentioned with different names in each paper, including their full name, abbreviation and various aliases. For example, 'support vector machine' is also called 'SVM', 'SVMs', 'support vector machines', 'support-vector machine', or 'support-vector machines'.

We summarized all of the names of an algorithm and compiled a dictionary of algorithms by removing duplicate words and manual classification. In the dictionary, each line represents an algorithm, and each line contains all of the names of the algorithms identified from ACL papers. Specifically, the dictionary includes 877 algorithms and 1,840 different names; examples of algorithms in the dictionary are displayed in Table 2.

Table 2. Examples about different names of algorithms

| Standard Name | Abbreviations and Aliases |
|---|---|
| Support Vector Machine | svm, svms, support vector machines, support-vector machines, support-vector machine |
| Conditional Random Field | crf, crfs, conditional random fields |
| Maximum Entropy | me, maxent, max-ent, maximum-entropy |
| Naive Bayes | nb, naïve bayesian, naivebayes |
| Expectation Maximization | em, em-algorithm, expectation-maximization, expectation and maximization, expectation-maximisation, expectation and maximisation, expectationmaximization |
| K Nearest Neighbor | knn, k-nn, k-nearest neighbor, k-nearest-neighbor, k-nearest neighbors, k-nearest-neighbors, k-nearest neighbour, k-nearest-neighbour, k-nearest neighbours, k-nearest-neighbours, k nearest-neighbor, k nearest neighbors, k nearest-neighbors, k nearest neighbour, k nearest-neighbour, k nearest neighbours, k nearest-neighbours |
| Context Free Grammar | cfg, cfgs, context free grammars, context-free grammar, context-free grammars, contextfree grammar, contextfree grammars |

### 3.4 Algorithm sentence extraction

Consider that the labeler may omit articles containing the algorithms in the dictionary, which may lead to inaccurate results. For example, the labeler found the *support vector machine* algorithm in 9 articles, but in fact, there were 10 articles mentioning *support vector machine*; we needed to find the article that was ignored by the labeler. To find all articles mentioning algorithms compiled in the dictionary, we matched the algorithm name with the full text of the academic paper in XML format. The algorithm sentences were also extracted; that is, as long as a name of an algorithm appeared in a sentence, we saved the ID of the article, the algorithm name and the algorithm sentence together. Through comparison with our manual annotation results, we could find the missing articles.

In addition, the algorithms have abbreviated names, and the same abbreviated name could represent different algorithm. For example, the BP algorithm can represent both the back-propagation algorithm and the belief propagation algorithm. There are also a few abbreviations that can represent both algorithms and other entities. For example, EM can represent the expectation maximization algorithm, or it may just represent the *m*-th entity.

Therefore, we disambiguated all of the algorithm sentences extracted by abbreviations. Specifically, if the full name and abbreviations of an algorithm appeared in an article simultaneously, then we assumed that the abbreviation in the text represented this algorithm. For example, the following two sentences were extracted from the same paper (Clark, 2002): "*This paper discusses the supervised learning of morphology using stochastic transducers, trained using the Expectation Maximization (EM) algorithm*" and "*We have presented some algorithms for the supervised learning of morphology using the EM algorithm applied to non-deterministic finite-state transducers.*" Since the first sentence gives the full name of the EM algorithm, we assumed that the EM in the second sentence indicated *expectation maximization.* If the full name did not appear in the article, we needed to make a further judgment based on the context of the algorithm sentence extracted by the abbreviation. In the final extraction result, the full name of all the abbreviations was added. Examples of matched sentences are shown in Table 3.

**Table 3. Example of sentence matched by the algorithm**

| ID | Algorithm | Full name | Matched Sentence |
|---|---|---|---|
| P01-1049 | SVM | Support vector machine | A large number of techniques have been proposed to tackle the problem, including regression model, nearest neighbor classification, Bayesian probabilistic model, decision tree, inductive rule learning, neural network, on-line learning, and, SVM (Yang & Liu, 1999; Tzeras & Hartmann, 1993). |

### 3.5 Analyzing the impact of algorithms

**(1) Indicator of influence**

In this paper, the number of papers mentioning an algorithm was utilized as an indicator to analyze the influence of the algorithm. Scholars have used various bibliometric indicators to evaluate the influence of papers (Cartes-Vellásquez & Manterola, 2014), authors (Fu & Ho, 2013) or institutions (Abramo et al., 2011), as well as knowledge entities. The number of mentions, citations, downloads, visits, and votes can be used as indicators to measure the impact (Ding et al., 2013; Urquhart & Dunn 2013; Howison et al., 2015; Pan et al., 2019). For the influence of the algorithm, the numbers of mentions (Wang & Zhang, 2018) and votes (Wu et al., 2008) were used to measure the influence in previous studies.

Compared with other indicators, the mention count is more suitable to measure the influence of an algorithm in academic papers. Because there are irregular citations to knowledge entities in academic papers (Mooney & Hailey, 2011), if we use the number of citations to measure the influence of algorithms, algorithms that are mentioned but not formally cited will be ignored; the mention count solves the problem. Furthermore, the counting unit of mentions could be either a sentence or an article. Considering the different writing styles of different authors, some authors repeatedly mention the algorithm in many sentences, and some authors may only mention the algorithm a few times; we choose the article as the counting unit to eliminate the inconsistency caused by different writing styles.

For the above reasons, in this article, the indicator of influence is the "mention count," and the counting unit is the "article."

**(2) Influence of different algorithms**

Because each algorithm has various names in different articles, we summarized the ID of articles that mention the same algorithm based on the dictionary and then removed the duplicate data. We used the number of papers that mention algorithms to evaluate the influence of algorithms. However, the total number of articles published each year is inconsistent, and the year in which each algorithm first appeared in ACL conference papers is also different; these cause interference in the measurement of the algorithm's influence. Therefore, on the basis of the number of papers mentioning algorithms, we also considered the total number of articles per year and the duration of influence. We first calculated the annual influence of the algorithm based on the publication time of papers; the sum of the annual influences was divided by the influence time of the algorithm, and finally the influence of algorithm *j* was obtained. Influence (*j*) is given by:

$$Influence(j) = \frac{\sum_{i=1979}^{2015} \frac{N_{ij}}{N_i}}{T_j} \qquad (1)$$

Where *i* represents the year, ranging between 1979 and 2015, $N_i$ is the total number of publications in year i, and $N_{ij}$ is the number of publications mentioning algorithm j in year *i*. Therefore, $\frac{N_{ij}}{N_i}$ is the *annual influence* of algorithm *j* in year *i*. $T_j$ is the duration from the year when algorithm *j* first appeared in ACL conference papers to 2015.

**(3) Different types of algorithms**

We further explored the impact of different categories of algorithms in ACL conference papers.

In most cases, algorithms in NLP articles are utilized to solve problems. By classifying the algorithms according to their functions, we can know what types of algorithms are mentioned in NLP papers, and then calculate the average influence of different types of algorithms. In this way, we can infer what the algorithm is mainly used to do in the NLP domain, and then judge what the common problems need to be solved in the NLP domain. In addition, we can study whether the influence of different algorithms in the same category affect each other.

Regarding the categories of algorithms, we referred to the category framework in Wikipedia and various textbooks (Christopher & Hinrich, 2000; Jennings & Wooldridge, 2012; Mitchell, 1997) and then proposed a preliminary classification framework. We invited experts in the field of NLP to optimize the framework. Finally, we divided the algorithms in the field of NLP into the following 14 types.

(i) Classification algorithm: a method that sorts data into labeled classes, or categories of information, on the basis of a training set of data containing observations whose category membership is known[4], for example, *support vector machine.*

*(ii)* Clustering algorithm: a method that groups a set of objects in such a way that objects in the same group are more similar to each other than to those in other groups[5], for example, *K-means.*

*(iii)* Dimension reduction algorithm: a method that reduces the number of random variables under consideration by obtaining a set of principal variables[6], for example, *singular-value decomposition.*

(iv) Grammar: a method that is used to represent the set of structural rules governing the composition of clauses, phrases and words in a natural language[7], for example, *context-free grammar.*

---

[4] https://deepai.org/machine-learning-glossary-and-terms/classifier
[5] http://en.volupedia.org/wiki/Cluster_analysis#Algorithms
[6] http://en.volupedia.org/wiki/ Dimension reduction
[7] http://en.volupedia.org/wiki/Grammar

*(v)* Ensemble learning algorithm: a method that uses multiple learning algorithms to obtain better predictive performance than could be obtained from any of the constituent learning algorithms alone[8], for example, *AdaBoost.*
   *(vi)* Link analysis algorithm: a data-analysis technique used to evaluate relationships (connections) between nodes[9], for example, *Pagerank.*
   *(vii)* Metric algorithm: an algorithm that defines the distance between each pair of elements of a set; they are usually utilized to evaluate the quality[10], importance, and similarity of texts, words, vectors and so on, for example, the *BLEU algorithm.*
   *(viii)* Neural networks: a computing method vaguely inspired by the biological neural networks that constitute animal brains, which learns to perform tasks by considering examples, generally without being programmed with task-specific rules[11], for example, *convolutional neural networks.*
   *(ix)* Optimization algorithm: a method that uses an iterative approach to make the result as close as possible to the optimal solution to the problem when the machine learning problem has no optimal solution or it is difficult to obtain the optimal solution[12], for example, *expectation maximization.*
   *(x)* Probabilistic graphical model: a probabilistic model for which a graph expresses the conditional dependence structure between random variables[13], for example, *the hidden markov model.*
   *(xi)* Regression algorithm: a method for estimating the relationships between a dependent variable and one or more independent variables[14], for example, *logistic regression.*
   *(xii)* Search algorithm: an algorithm that solves a search problem, namely, to retrieve information stored within some data structure or calculated in the search space of a problem domain, either with discrete or continuous values[15], for example, the *beam-search algorithm.*
   *(xiii)* Unique algorithm in the NLP domain: an algorithm only used to deal with natural language processing tasks. For example, the *CKY algorithm.*
   *(xiv)* Other: an algorithm that cannot be divided into 13 categories mentioned above, for example, the *smith-waterman algorithm.*
   Using this framework, we classify algorithms and explore the types of algorithms with high influence.

**(4) Influence of algorithms in different years**
According to the calculation results of each algorithm's annual influence in different years, we ranked the influence of the algorithm each year and discussed the differences among the algorithms with high influence.

**(5) Evolution of the influence of different algorithms**
Based on the influence of each algorithm ach year, we took each algorithm as the research object and analyzed the changes in the annual influence of each algorithm over time. After

---

[8] http://en.volupedia.org/wiki/Ensemble_learning
[9] http://en.volupedia.org/wiki/Link_analysis
[10] http://en.volupedia.org/wiki/Metric_(mathematics)
[11] http://en.volupedia.org/wiki/Artificial_neural_network
[12] http://en.volupedia.org/wiki/Mathematical_optimization#Optimization_algorithms
[13] http://en.volupedia.org/wiki/Graphical_model
[14] http://en.volupedia.org/wiki/Regression_analysis
[15] http://en.volupedia.org/wiki/Search_algorithm

that, algorithms are classified according to different trends in evolution.

## 4. Results

As described in Section 3.5, we obtain the influence of algorithms in the NLP domain, influence of different types of algorithms, and evolution of the algorithms. The results are presented in this section.

### 4.1 Influence of different algorithms in ACL papers

We answer *RQ1* in this section. We collected 4,641 papers, of which 4,043 papers mentioned algorithms, accounting for 87% of all papers. The result shows that algorithm entities are widely studied in the NLP field and that they play an important role.

**(1) Top-10 algorithms with the highest influence**

The influence of different algorithms is calculated by the formula given in Section 3.5 (1). Due to the space limitation, this section only shows the top-10 algorithms with the highest influence. As displayed in Table 4, most of these algorithms are classical and basic algorithms. The support vector machine (SVM) takes the first place with the absolute advantage in quantity. As described in *The Top Ten Algorithms in Data Mining* (Wu et al., 2008), the SVM algorithm has a solid theoretical foundation and is one of the most stable and accurate algorithms among all well-known algorithms. What comes next is context-free grammar; this algorithm is powerful enough to express the syntax pattern of most programming languages. Besides, context-free grammar is simple enough for scholars to construct an effective analysis algorithm to determine whether a given string is generated by a context-free grammar (Kadlec, 2008). The BLEU algorithm, namely, the bilingual evaluation replacement algorithm, ranks third in the list. BLEU is one of the first metrics to claim a high correlation with human judgments of quality and remains one of the most popular automated and inexpensive metrics (Papineni et al., 2002). Maximum entropy comes next, which is a classical algorithm for solving classification problems.

Among these 10 algorithms, the most special is word2vec. As an emerging word vector representation method, it combines the advantages of high accuracy and low computational cost (Mikolov et al., 2013). It is not a classic algorithm with a long history, but it has also gained high influence in ACL papers, showing its superior performance. In general, support vector machine has the highest influence in the field of NLP. Most of the high-impact algorithms are classic algorithms, but the influence of emerging algorithms cannot be ignored either.

**Table 4. The top-10 most influential algorithms in ACL conference papers**

| Rank | Algorithm | Rank | Algorithm |
|---|---|---|---|
| 1 | Support vector machine | 2 | Context-free grammar |
| 3 | BLEU algorithm | 4 | Maximum entropy |
| 5 | Word2vec | 6 | Expectation maximization |
| 7 | Hidden markov model | 8 | Dynamic programming algorithm |
| 9 | Decision-tree | 10 | Cosine similarity |

**(2) The type of high-impact algorithms**

Table 5. The type of high-impact algorithms

| Rank | Type | Number of algorithms | Average influence |
|---|---|---|---|
| 1 | Classification algorithm | 15 | 0.0448 |
| 2 | Probabilistic graphical models | 7 | 0.0410 |
| 3 | Grammar | 14 | 0.0400 |
| 4 | Optimization algorithm | 16 | 0.0382 |
| 5 | Neural networks | 10 | 0.0364 |
| 6 | Ensemble learning algorithm | 2 | 0.0354 |
| 7 | Metric algorithm | 11 | 0.0342 |
| 8 | Unique algorithms in NLP domain | 4 | 0.0327 |
| 9 | Regression algorithm | 2 | 0.0264 |
| 10 | Search algorithm | 3 | 0.0240 |
| 11 | Dimension reduction algorithm | 1 | 0.0233 |
| 12 | Other | 9 | 0.0233 |
| 13 | Link analysis algorithms | 2 | 0.0222 |
| 14 | Clustering algorithm | 4 | 0.0146 |

Because the task of identifying categories of all algorithms is time-consuming, we obtain the top-100 algorithms according to the influence and manually classify them according to the classification framework introduced in Section 3.5. A professor who are familiar with natural language processing reviewed and improved the classification results. Then we counted the number of algorithms in each category and calculated the average influence of each category. The results are shown in Table 5.

Among the top-100 algorithms, the average influence of the classification algorithm ranks the first, and the number of algorithms in this category ranks the second, which means that the influence of classification algorithms is significantly higher than other categories. Classification is a basic task in the field of NLP, it can be regarded as a subtask in different research topics, and for instance, recognizing named entities with the help of classification algorithms and sequence labeling models. There are also separate classification tasks in NLP, including citation intent classification, text classification, emotion classification, etc. In contrast, although optimization algorithms make up the largest proportion in the top-100 list, their average influence only ranks the fourth. We can speculate that this is because optimization is often solely a part or a step of a NLP task, which makes the result be as close to the optimal solution as possible. Apart from those, the probability graph model demonstrates the feature of "small but excellent", and a small number of algorithms achieve high average influence. However, the proportions of link analysis algorithms and regression algorithms in the list of high-impact algorithms are relatively small.

**4.2 Top-10 Algorithms in Different Ages**

We answer *RQ2* in this section. Considering the space limitations, we provide a list of top-10 algorithms with high influence each year. As displayed in Appendix 1, it could be easily found out that the differences between popular algorithms in different years. Generally, the high-impact algorithms in different generations reveal the following trends: the early high-impact algorithm was the syntax analysis algorithm, which later became the traditional machine learning algorithm. What is more, deep learning algorithms had the highest influence in the last two years. Dynamic demonstration of algorithm influence in NLP can be accessed at https://public.flourish.studio/visualisation/3073870/.

Based on the Appendix 1, we counted the number of algorithms in the top-10 list grouped by three types to explore the evolution of high-impact algorithms in different eras. As shown in Figure 3, the changes in the number of high-impact syntactic analysis algorithms and machine learning algorithms demonstrate completely opposite trends. To be specific, the first period of dramatic changes appeared between 1992 and 1994, during which the number of machine learning algorithms increased significantly, while the syntactic analysis algorithms showed the opposite. After that, in 1997, the number of high-impact machine learning algorithms surpassed syntactic analysis algorithms for the first time. The second period of dramatic changes was from 2001 to 2002, during which the gap between the numbers of the two became larger. Since 2013, the proportion of machine learning algorithms in the top-10 has begun to decline. On the contrary, the share of deep learning algorithms in the high-impact algorithm list was increased. According to the results listed in Figure 3, we identify three major epochs in the algorithm history in NLP: The syntax analysis period (1979-1996), traditional machine learning period (1997-2013), the deep learning period (2014 to present).

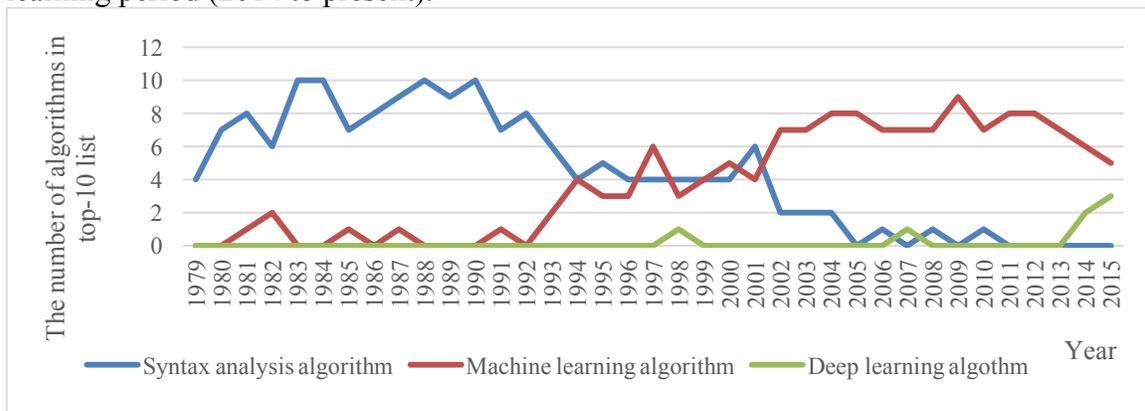

**Figure 3. The number of papers accepted by ACL conference each year**

Based on Appendix 1, we selected a representative algorithm from each period, which includes context free grammar, support vector machine and neural network. Figure 4 shows the changes in their rankings in the top ten each year. According to the ranking of these algorithms, we analyze the characteristics of algorithms each stage in detail in the following text.

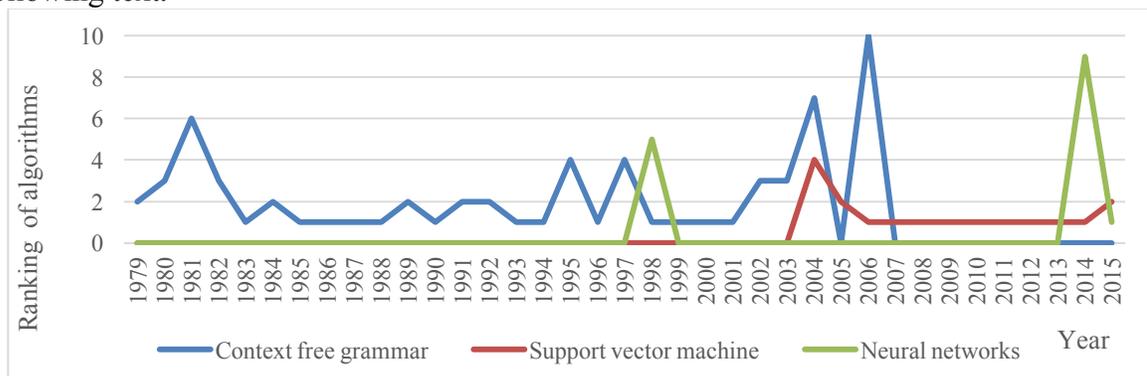

**Figure 4. Ranking of representative algorithms in the top-10 each year**

In the period of syntactic analysis, more than half of the 10 algorithms were grammars, and the influence of context-free grammar (CFG) was particularly excellent. According to Appendix 1 and Figure 4, we found that during the 18 years from 1979 to 1996, CFG appeared annually and won the first place for nine years. Most of these influential algorithms are used for syntactic analysis, and other algorithms are related to syntactic analysis. In addition to the popular syntax analysis algorithm, such as Cocke-Younger-Kasami algorithm and the Earley algorithm, and the Hobbs algorithm is a coreference resolution algorithm based on the syntactic analysis, which indicated that scholars were focusing on linguistic research during this period.

In the period of machine learning, the important role of machine learning algorithms in ACL papers was revealed. Although the hidden markov model and the decision tree appeared in the top-10 list before, the rankings were relatively low. It was not until 1997 that the decision tree occupied the first place. Between 1997 and 2001, syntactic analysis algorithms and machine learning algorithms took up more than half of the position in turn, and CFG still often appeared in the first place. After 2001, an increasing number of machine learning algorithms entered the top-10 list, while the proportion of parsing algorithms began to decrease sharply. In 2006, support vector machines had the highest influence remaining the first until 2014. Because, at this stage, large-scale text processing has become the main target of natural language processing, and scholars were increasingly adopting automatic machine learning algorithms to acquire language knowledge.

In the period of deep learning, a growing number of deep learning algorithms appeared in the top-10 list and occupied an important position. However, the number of traditional machine learning algorithms in the high-impact list was decreased. In 2015, neural networks took the first place, and word2vec and Skip-gram also appeared in the top-10 list. Admittedly, traditional machine learning algorithms rely on large-scale manual corpus tagging and feature engineering. However, constructing effective labeled corpus and classification features are still characterized with being time-consuming and labor-intensive. Compared with the traditional algorithms, the deep learning algorithm reduces manual investment. On the one hand, it solves the problem of data sparseness caused by the high-dimensional vector space. On the other hand, the word vector contains more semantic information than the manually selected features. Therefore, the influence of deep learning algorithms will gradually exceed that of traditional machine learning algorithms. It is believed that this trend will become more obvious in the future.

**4.3 The evolution of influence of different algorithms**

We answer *RQ3* in this section. For each algorithm, we have drawn trend graphs for the evolution of influence over time, and algorithms with similar trends of evolution were analyzed together. Three types of change trends are shown in Figure 3, Figure 4 and Figure 5.

**(1) Algorithms with rapidly growing influence**
The first type is called sharp growth. Nine algorithms that conform to this trend were selected. As shown in Figure 3, the influence of these algorithms increased rapidly in a certain year.

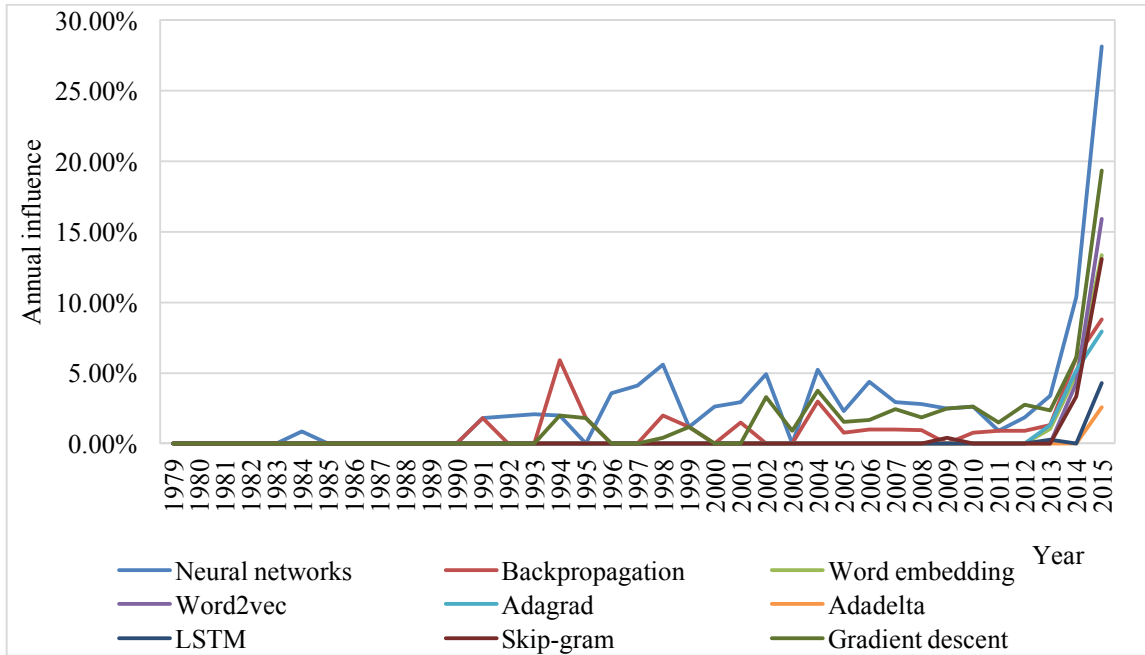

**Figure 5. Algorithms with rapidly growing influence**

These algorithms can be subdivided into two subcategories. One subtype is the algorithm for which the influence in the early years was not high but began to grow rapidly after one year. In Figure 5, algorithms in line with this trend are back-propagation, Brown clustering, neural networks, etc. They are not newly proposed algorithms and were mentioned in early years. It is deemed that at a certain time, scholars found that these algorithms could be utilized to solve some new tasks, or broke the technical restrictions that had previously restricted their development.

Taking the neural network model as an example, in our results, the first peak of influence appeared in 1984, which then entered the stable period after the 1990s, but the influence began to grow rapidly after 2011. The neural network was born in the 1940s, and gained great popularity in the 1980s. However, after the 1990s, its development was less outstanding. On the one hand, the rise of statistical learning methods suppressed the development of neural networks. On the other hand, as the number of neural network layers increases, the difficulty of training the neural network increases geometrically, and the lack of computing resources once again limits the development of the model. Fortunately, in 2006, Hinton and Salakhutdinov (2006) solved the problem of how to set initial values in neural network learning and proposed a method for quickly training deep neural networks. After that, the neural network model entered a prosperous period after 2010. According the results, the influence of the neural network model also entered a growth period after 2010 and began to grow strikingly in 2013. The main reason is that the training methods of neural network models have been improved. At the same time, with the development of science and technology, scholars obtained powerful computing resources to train the model.

Another subtype is algorithms that had no article mentioning them in the early years, but after their first mention, their influence showed a sharp increase. These algorithms include word2vec, AdaGrad and Skip-gram. Obviously, they are all newly proposed methods. Taking the AdaGrad algorithm as an example, as one of the most commonly used optimization algorithms in deep learning, it was put forward in 2010. It adjusts the learning rate of each dimension and avoids the problem in which the unified learning rate cannot adapt to all dimensions (Duchi et al., 2011). In general, these algorithms are mostly adapted to solve deep learning tasks, so it is not difficult to speculate on the reason for rapid growth of influence; compared to traditional statistical machine learning algorithms, deep learning algorithms have achieved better performance in natural language processing in recent years.

**(2) Algorithms with steadily growing influence**

The second type is called stable growth. As displayed in Figure 6, for algorithms in this type, they have been mentioned for a long time, and from the first year when the algorithm was mentioned, their influence has fluctuated over time, but the overall evolution has shown an upward tendency. All algorithms can be called classic algorithms, and the results in Figure 6 demonstrates that a classical algorithm can pass the test of time and prove its value. Regardless of changes over time and whether new algorithms are developed, the influence of classic algorithms maintains stable growth. They usually have features that cannot be replaced by new methods, which may be due to the fact that they are simpler to use or have lower cost, and undoubtedly, they are more likely to be compared with new methods as a baseline. In general, they give rise to more far-reaching effects.

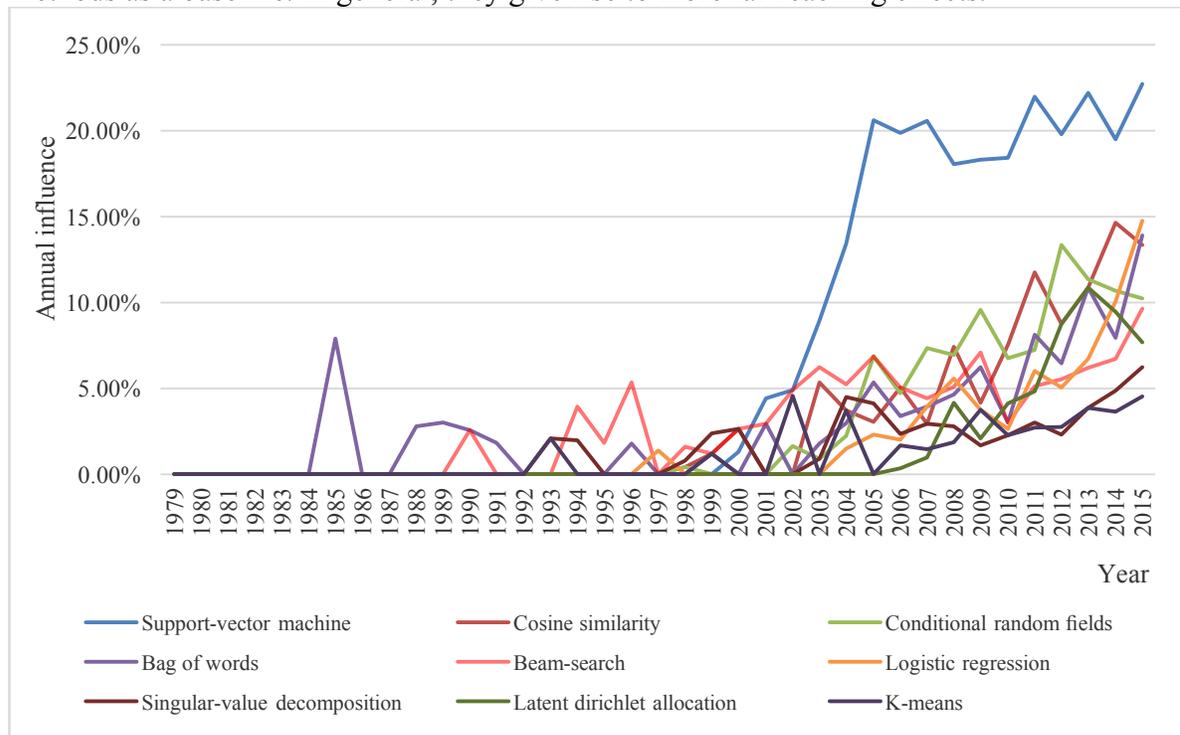

**Figure 6. Algorithms with steadily growing influence**

Among the nine algorithms, the influence of support vector machines shows the highest growth, which once again proves our results in 4.1. Support vector machine is indeed a very influential algorithm whose influence continues to increase. In general, influence of the nine algorithms appeared in the 1980s and began to show a clear growing tendency approximately around 2000. As the processing speed and storage capacity of computers have increased significantly since the mid-1990s, statistical machine learning algorithms have developed speedily since the 1990s. On the other hand, during the 1990s, due to the commercialization of the Internet and the development of network technology, the need for information retrieval and information extraction based on natural language became more prominent, and the influence of the algorithms used to process these two tasks, for example the Beam-search, naturally ascended.

**(3) Algorithms with steadily declining influence**

The third type is called stable decline. The influence of an algorithm in this type shows a downward trend over time. For algorithms in this type, it is speculated that, with the development of new algorithms, the performance or application of these algorithms no longer presents advantages, and scholars have more new choices for their own research.

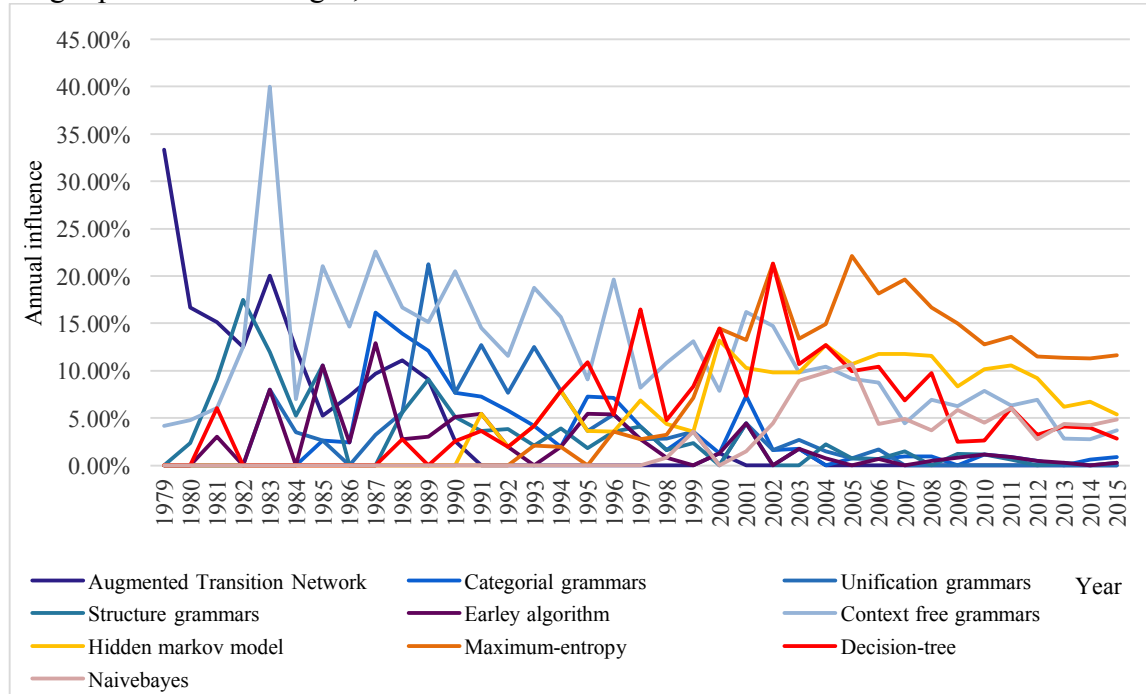

**Figure 7. Algorithms with steadily declining influence**

As shown in Figure 7, there are two types of algorithms showing a declining trend of influence. The first type is the grammar whose influence before 1990 was significantly higher than that after 1990, and the influence became very low after 2000. In contrast, according to our previous analysis, the influence of machine learning algorithms was increasing during this period. Among these algorithms, the showpieces include the augmented transition network (transition network grammar) and the context-free grammar. The influence of the augmented transition network has been declining since the first year (1979). Although the ATN (Woods, 1970) has been widely used in the research of human-computer conversation and machine translation, the algorithm relies too much on grammar, which makes it unable to deal with sentences that go against the rules of grammar. When processing flexible objects, it cannot meet the needs. The context-free grammar represents an algorithm for which the influence increase first and then decrease. Because the context-free grammar can easily deduce the grammatical structure of a sentence, but the generated structure may be ambiguous.

The second type is traditional machine learning algorithms. Before 2002, their influence showed an upward trend. Between 2002 and 2005, they entered a relatively flat stage of development, and then began to decline over time. Based on the tendency, it is not difficult to judge that the increase in its influence is related to the popularity of machine learning algorithms in the NLP field, but the decline of influence is due to deficiencies of different algorithms. For example, the decision tree is simple to understand and to interpret, but it can be unstable because small variations in the data might result in a completely different tree[16], and when there are more categories, errors may increase correspondingly. Naïve bayes can handle multi-classification tasks with higher stability, but it requires the independence of data set attributes. Otherwise, the classification performance is projected to be unsatisfactory.

In general, the influence of the above two algorithms decreases with time going by, but they show a clear cut-off point. Before 1996, grammar was more influential than machine learning algorithms. After 1996, the situation was just the opposite, which was also in line with the conclusion we reached in Section 4.2. Therefore, no matter what it analyzes, the influence of syntactic analysis algorithms is gradually being replaced by machine learning algorithms.

According to the previous research results, although the comprehensive influence of the grammar is relatively high, its overall influence shows a downward trend. Currently, the construction of various corpora indicates that people pay more attention to the processing of large-scale real texts, which are difficult to process by using the traditional rule-based parsing technology. Taking information extraction as an example, when the amount of end-to-end data is sufficient to extract information directly, scholars are able to achieve the goal directly by using machine learning algorithms and no longer need the syntax analysis. Therefore, the influence of machine learning algorithms increased from 1990 to 2000, while the influence of grammars in the field of NLP began to decrease year by year.

**(4) Time span from appearance to height**

Concerning the 28 algorithms mentioned above, we further analyze the time from their emergence to the highest influence, which is called the rising span. Displayed in Table 6, the rising span of algorithms shows different patterns. For algorithms with growing influence, most of them got the highest influence in 2015. The rising span of the algorithm with rapidly growing influence is mostly lower than that of the algorithm with steadily growing influence, the former is often less than five years, and the latter is almost more than 15 years. The longest can reach as long as 31 years. For algorithms with steadily decreasing influence, the rising span is almost 10 years. Comparing the two kinds of algorithms with stable influence, it can be found that the rising span of the influence reduction algorithm was shorter, although many of them appeared in ACL earlier, which means that, in the long run, people can predict its development trend in a shorter time.

Additionally, we find that different algorithms with the same function show progressive trend in influence changes. Taking the classification algorithm as an example, the classification algorithms in Table 6 include support vector machines, the decision tree and naïve bayes. The decision tree appeared and reached the peak of influence earliest, however, its influence began to decline significantly after 2002, while the influence of naive Bayes still maintained the good momentum of development, and surpassed the decision tree in 2005. In contrast, the support vector machine (SVM) appeared when influence of the two algorithms was close to the peak, and entered the popular period in 2003, which was the year when the influence of the former two algorithms began to decline. Similarly, the probability graph model shows the same trend. When the influence of hidden markov model (HMM) showed a downward trend, conditional random fields (CRF) just appeared the ACL papers, and began to enter a stable growth period. As classical probabilistic graph models, they actually play an important role in sequence annotation tasks. However, compared with HMM, CRF can provide more reasonable probability normalization results and global optimal solutions, so it is not difficult to explain why its influence shows the more obvious growth.

---

[16] https://dirtysalt.github.io/sklearn.html

Table 6. Rising span of different algorithms

| | Algorithms | The year of first appearance | The year of highest impact | Time span |
|---|---|---|---|---|
| **Algorithms with rapidly growing influence** | Neural networks | 1984 | 2015 | 31 |
| | Back propagation | 1991 | 2015 | 24 |
| | Word embedding | 2013 | 2015 | 2 |
| | Word2vec | 2013 | 2015 | 2 |
| | Adagrad | 2012 | 2015 | 3 |
| | Adadelta | 2015 | 2015 | 0 |
| | LSTM | 2013 | 2015 | 2 |
| | Skip-gram | 2009 | 2015 | 6 |
| | Gradient descent | 1994 | 2015 | 21 |
| **Algorithms with steadily growing influence** | Support-vector machine | 2000 | 2015 | 15 |
| | Cosine similarity | 1998 | 2014 | 16 |
| | Conditional random fields | 1998 | 2012 | 14 |
| | Bag of words | 1985 | 2015 | 30 |
| | Beam-search | 1990 | 2015 | 25 |
| | Logistic regression | 1997 | 2015 | 18 |
| | Singular-value decomposition | 1993 | 2015 | 22 |
| | Latent dirichlet allocation | 2006 | 2013 | 7 |
| | K-means | 1993 | 2015 | 22 |
| **Algorithms with steadily declining influence** | Augmented transition network | 1979 | 1979 | 0 |
| | Categorial grammars | 1985 | 1987 | 2 |
| | Unification grammars | 1983 | 1989 | 6 |
| | Structure grammars | 1980 | 1982 | 2 |
| | Earley | 1981 | 1987 | 6 |
| | Context free grammars | 1979 | 1983 | 4 |
| | Hidden markov model | 1991 | 2000 | 9 |
| | Maximum entropy | 1993 | 2005 | 12 |
| | Decision tree | 1983 | 2002 | 19 |
| | Naïve bayes | 1998 | 2005 | 7 |

## 5. Discussion

### 5.1 Algorithm evolution and domain evolution

Cambria & White (2014) reviewed the development of the NLP field in his research. They pointed out that most NLP research carried out in the early days focused on syntax analysis and statistical NLP has been the mainstream NLP research direction since the late 1990s. According to our research, on the basis of the development of algorithms, research in the field of NLP has indeed shown this trend. In the early ACL papers (before 1996), algorithms that accounted for greater ratios were various grammars. On the one hand, in the field of computational linguistics, there may be more research studies on linguistics in the early stage, and grammar is indeed a necessary method. On the other hand, some scholars believe that syntactic processing is necessary in many NLP tasks. Therefore results of this article reveal that some grammars, e.g., context-free grammar, still plays an important role in the top-10 list after 2000. The so-called syntax-driven characteristic refers to the strategy adopted by researchers in their work to first solve the grammar, which makes machine learning techniques more directly applicable. After grammar was widely mentioned, in the late 1990s, statistical machine learning algorithms began to appear in the top-10 list and occupied a dominant position after 2005. In recent years, deep learning algorithms (e.g. neural networks) got higher impact. It should be noted that because we were only able to obtain the data before 2016, if the papers published after 2015 could be obtained, the tendency of deep learning algorithms to dominate would become more apparent. Comparing our preliminary results and experts' reviews, we can find that changes in the influence of the algorithm can also reflect the development of the research field. NLP has witnessed the emergence of several subfields, from the early grammar-based approaches in the 1950s-1970s, to the statistical revolution in the 1990s, to the recent deep learning algorithms (Jurgens et al., 2018).

**5.2 Reasons for changes in the influence of algorithms**

We analyzed the different modes of changes in the influence of algorithms in section 4.3. It is obvious that the influence of some algorithms increases with time and even increases rapidly in a short time, while the influence of some algorithms gradually decreases or even disappears. We think that the reasons for the change in influence can be summarized by the following three points.

**(1) The performance of the algorithm itself**

For algorithms that have high influence and an overall upward trend of influence, they usually have some excellent characteristics, for example, a solid theoretical basis, wider range of application, high stability, and low resource consumption. Support vector machine is representative of such algorithms. Algorithms with reduced influence are often easily replaced by new algorithms because of poor performance.

**(2) Development of technology**

In the early days, due to the immaturity of technologies, some algorithms encountered a bottleneck period during the development, limiting their early influence. However, with the development of science and technology, the computing and storage capabilities of computers have been greatly improved, an increasing number of data resources are available to people, and the technical and resource problems that previously limited the performance of algorithms have been solved. Therefore, some algorithms are widely used again, and their influence has increased significantly.

**(3) Changes in user demands**

After 1990, in the field of NLP, scholars needed algorithms to automatically process larger-scale data. The development of the Internet has also greatly increased the demand for information mining and information retrieval algorithms. Traditional rule algorithms can no longer meet these needs. Thus, in our results, the influence of grammar is gradually replaced by the influence of machine learning algorithms. It is at this stage that statistical revolution takes place (Jurafsky, 2015).

**5.3 The top-ten data mining algorithms in the NLP domain**

Wang and Zhang (2018) used the numbers of papers that mentioned algorithms to conduct a preliminary inquiry into the influence of the top ten data mining algorithms in the NLP field. The results are shown in Table 7. As classic methods of data mining, the 10 algorithms are all famous, but their influence in the field of NLP exhibits a difference.

When we compare other algorithms with these ten algorithms, the advantages of data mining algorithms in the NLP field are not obvious. Although SVM and EM algorithms still appear in the top-10 algorithm list, the remaining algorithms are no longer data mining algorithms. In our results, among the 10 most influential algorithms in the ACL papers, in addition to data mining algorithms, there are also grammar and statistical learning algorithms. It can be speculated that classic data mining algorithms have their advantages, but not all classic data mining algorithms could get high influence in different fields. The influence of the algorithm in a particular domain will inevitably present its own characteristics.

Table 7. The influence of top-10 date mining algorithms in NLP domain

| No. | Algorithm | # papers (Ratio) | No. | Algorithm | # papers (Ratio) |
|---|---|---|---|---|---|
| 1 | SVM | 774 (44.33%) | 6 | KNN | 67 (3.84%) |
| 2 | EM | 403 (23.0%) | 7 | C4.5 | 57 (3.26%) |
| 3 | Naive Bayes | 190 (10.88%) | 8 | AdaBoost | 21 (1.20%) |
| 4 | K-Means | 115 (6.59%) | 9 | Apriori | 14 (0.80%) |
| 5 | PageRank | 92 (5.27%) | 10 | CART | 13 (0.74%) |

**5.4 Differences between our method and other methods**

In the existing work on algorithm influence measurement, the representative method is expert voting and the number of mentions. For the expert voting, in September 2006, the ICDM hosted a poll of classic algorithms in data mining. The sponsor invited various experts to recommend candidate algorithms and then evaluated the influence of candidate algorithms through the expert voting. However, there are some limitations. First, this method of selecting and evaluating algorithms relies on the experts' personal experience and ideas and lacks detailed quantitative data to support the result. Second, organizing expert voting is very time-consuming, and such events cannot be held frequently. Third, the results did not examine the influence of algorithms in a specific field and could not provide fundamental data for subsequent applications. Regarding number of mentions, we have introduced Wang's work in section 5.3. Although their work also uses mention count to evaluate the influence of the algorithms, they only explored the influence of ten algorithms. In addition, they directly used the number of papers to evaluate the influence of the algorithms within a period without considering the impact of the difference in the total number of articles published each year on the numbers of articles that mentioned the algorithms. Different from the above research, this article first uses manual annotation to collect algorithm entities from academic papers. Compared with expert recommendations, we can obtain more objects. Subsequently, we use the number of articles to evaluate the impact of the algorithms and consider the total number of articles each year and the duration of influence, which is more convenient than inviting experts and ensuring the objectivity of the results.

# 6. Conclusion and Future Works

To explore the algorithms in the academic papers in a domain, this paper takes the NLP domain as an example and identifies algorithms from the full-text content of papers by manual annotation. Influence of algorithms is analyzed based on the number of articles. It should be noted that although this paper focuses on algorithms in the field of NLP, the methodology can be applied to identify algorithms in other fields or disciplines. Our results show that among different algorithms in the NLP domain, SVM has the highest influence due to its stability and accuracy. At the same time, for different types of algorithms, the influence of optimization algorithms is significantly higher than that of other categories. The change in algorithm influence over time also reflects the development of the research field; in the early stage, the popular algorithms were grammar in linguistic research, and then it gradually became statistical learning algorithms. In recent years, deep learning algorithms begun to occupy the dominant position. Among the evolution of all kinds of influence, there are three obvious trends, namely, the steady growth of classical algorithms, the rapid growth of new algorithms, and the steady decline of algorithms.

The contribution of this article is threefold. First, for the first time, this paper identifies more algorithms mentioned in papers in a specific field, not only those with pseudocode or detailed descriptions (Tuarob et al., 2016). Common algorithms are not omitted because many algorithms mentioned in papers are collected. Compared with the traditional bibliographic information and citation content, the full-text content provides more algorithms that are mentioned but not cited in academic papers. Second, we discuss the influence of algorithms from various perspectives. In addition to the commonly mentioned algorithms, we also analyze the evolution of algorithm influence. Third, this work provides more objective results regarding the influence of algorithms. This work collects more authors' opinions and directly identifies algorithms from their papers, and it evaluates influence based on the number of papers and duration of influence simultaneously.

As a preliminary study, this paper has some limitations. First, we annotate the algorithms in articles manually; although manual tagging can identify most of the algorithms in the article, we cannot guarantee that all of the algorithms are identified. We use the method of dictionary matching to ensure that all articles that mention the algorithm in the dictionary can be found, but if the algorithm is not included in the dictionary at the beginning, we cannot get the relevant sentences. However, we do not believe that the missing algorithms have a negative impact on our results, and the impact of missing annotations on our current results is not significant, because the algorithms with high influence; that is, the algorithms often mentioned in the papers would not be ignored by the annotator. Second, in this paper, the indicator of influence is only the number of articles mentioning algorithms. Other syntactic features, such as chapters, and semantic features, are not considered to measure the influence. Finally, we only collect the full text of conference papers in the ACL meeting to explore algorithms concerning NLP. There are other conference papers and journal papers in the field of natural language processing in the ACL anthology （http://www.aclweb.org/anthology/）, but they do not provide the full-text content in XML format. The ACL reference corpus provides a dataset in XML format concerning other conference papers, but it only gives articles published over a few years. As a result, although we can manually extract the algorithms from more articles in PDF format, we cannot retrieve them again in the full text in XML format to ensure that we can collect all of the articles mentioning each algorithm (work in section 3.4).

In the future, we will collect more conference papers and journal papers in the field of NLP to ensure that the dataset can cover more NLP research and attempt to transform them into structured data that machines can process. Subsequently, we intend to utilize the results of this article as training data, which means that algorithms and algorithm sentences in this work will be used to train machine learning models, and an optimal model based on performance will be selected to carry out the automatic extraction of algorithm entities. Furthermore, we will apply the full-text content to explore other features of the algorithms, such as the location function and relationship of the algorithms, and use a variety of features to conduct a comprehensive evaluation of the extracted algorithms. Finally, we intend to explore the specific tasks solved in the article using the algorithm so that we can not only understand the reason for the change of the algorithm's influence from the perspective of the task but also recommend classic and emerging algorithms according to the task. The results of the influence analysis of the algorithms can provide recommendations for scholars.

## Acknowledgements

This study is supported by the National Natural Science Foundation of China (Grant No. 72074113), Science Fund for Cre-ative Research Group of the National Natural Science Foundation of China (Grant No. 71921002) and Postgraduate Research& Practice Innovation Program of Jiangsu Province (Grant No. KYCX19 0347).

# Appendix 1. Top-10 algorithms with higher influence in each year

| Year | 1 | 2 | 3 | 4 | 5 | 6 | 7 | 8 | 9 | 10 |
|---|---|---|---|---|---|---|---|---|---|---|
| 1979 | Augmented transition network | Context free grammars | Hobbs algorithm | Kernel methods | Case frame | Local maxima | Discrimination network | * | * | * |
| 1980 | Augmented transition network | Hobbs algorithm | Context free grammars | Semantic grammars | Case frame | Fuzzy matching | Higher order unification | Structure grammars | Cocke younger kasami | Plan recognition |
| 1981 | Augmented transition network | Structure grammars | Lexical functional grammars | Left corner | Hobbs algorithm | Context free grammars | Decision tree | Semantic grammars | Case frame | Cocke younger kasami |
| 1982 | Structure grammars | Augmented transition network | Context free grammars | Semantic grammars | Plan recognition | Classification and regression tree | Dynamic programming | Hyperlink induced topic search | Generalized phrase structure grammars | Lexical functional grammars |
| 1983 | Context free grammars | Augmented transition network | Lexical functional grammars | Structure grammars | Tree adjoining grammars | Definite clause grammars | Semantic grammars | Earley algorithm | Unification grammars | Hobbs algorithm |
| 1984 | Augmented transition network | Context free grammars | Lexical functional grammars | Structure grammars | Case frame | Definite clause grammars | Unification grammars | Semantic grammars | Dependency grammars | Functional unification grammars |
| 1985 | Context free grammars | Lexical functional grammars | Structure grammars | Definite clause grammars | Earley algorithm | Head grammars | Plan recognition | Left corner | Head drive phrase structure grammar | Bag of words |
| 1986 | Context free grammars | Tree adjoining grammars | Generalized phrase structure grammars | Lexical functional grammars | Augmented transition network | Kimmo generation | Definite clause grammars | Head grammars | Earley algorithm | Plan recognition |
| 1987 | Context free grammars | Categorial grammars | Earley algorithm | Augmented transition network | Definite clause grammars | Tree adjoining grammars | Lexical functional grammars | Head grammars | Functional unification grammars | Constraint propagation |
| 1988 | Context free grammars | Categorial grammars | Augmented transition network | Tree adjoining grammars | Head drive phrase structure grammar | Unification grammars | Categorial unification grammar | Linear indexed grammar | Earley algorithm | Definite clause grammars |
| 1989 | Unification grammars | Context free grammars | Categorial grammars | Head drive phrase structure grammar | Unification categorial grammars | Augmented transition network | Tree adjoining grammars | Structure grammars | Lexical functional grammars | Centering algorithm |
| 1990 | Context free grammars | Tree adjoining grammars | Combinatory categorial grammars | Unification grammars | Categorial grammars | Head drive phrase structure grammar | Functional unification grammars | Cocke younger kasami | Lexical functional grammars | Categorial unification grammar |
| 1991 | Head drive phrase structure grammar | Context free grammars | Unification grammars | Tree adjoining grammars | Categorial grammars | Dynamic programming | Plan recognition | Earley algorithm | Centering algorithm | Hidden markov model |
| 1992 | Tree adjoining grammars | Context free grammars | Unification grammars | Combinatory categorial grammars | Left corner | Categorial grammars | Plan recognition | Functional unification grammars | Structure grammars | Lexical functional grammars |
| 1993 | Context free grammars | Unification grammars | Tree adjoining grammars | Head drive phrase structure grammar | Expectation maximization | Centering algorithm | Left corner | Categorial grammars | Cocke younger kasami | Hidden markov model |
| 1994 | Context free grammars | Head drive phrase structure grammar | Left corner | Dynamic programming | Unification grammars | Expectation maximization | Hidden markov model | Decision tree | Maximal likelihood estimation | Head grammars |
| 1995 | Tree adjoining grammars | Head drive phrase structure grammar | Decision tree | Context free grammars | Expectation maximization | Combinatory categorial grammars | Categorial grammars | Yarowsky | Earley algorithm | Centering algorithm |
| 1996 | Context free grammars | Tree adjoining grammars | Dynamic programming | Head drive phrase structure grammar | Maximal likelihood estimation | Kaplan and kay | Categorial grammars | Yarowsky | Decision tree | Earley algorithm |
| 1997 | Decision tree | Expectation maximization | Maximal likelihood estimation | Context free grammars | Dynamic programming | Head drive phrase structure grammar | Tree adjoining grammars | Hidden markov model | Definite clause grammars | Binary trees |
| 1998 | Context free grammars | Head drive phrase structure grammar | Dynamic programming | Tree adjoining grammars | Neural networks | Decision tree | Dependency grammars | Hidden markov model | Yarowsky | N best |

| Year | | | | | | | | | |
|------|---|---|---|---|---|---|---|---|---|
| 1999 | Context free grammars | Probabilistic context free grammar | Dynamic programming | Decision tree | Maximum entropy | Head drive phrase structure grammar | Tree adjoining grammars | N best | C4.5 | Left corner |
| 2000 | Decision tree | Maximum entropy | Hidden markov model | Maximal likelihood estimation | Expectation maximization | Context free grammars | Head drive phrase structure grammar | Tree adjoining grammars | Dependency grammars | Edit distance |
| 2001 | Context free grammars | Maximum entropy | Dynamic programming | Hidden markov model | Head drive phrase structure grammar | Tree adjoining grammars | Decision tree | Dependency grammars | Probabilistic context free grammar | Categorial grammars |
| 2002 | Maximum entropy | Decision tree | Context free grammars | Expectation maximization | Probabilistic context free grammar | Yarowsky | Dynamic programming | Hidden markov model | N best | Naive Bayes |
| 2003 | Maximum entropy | Decision tree | Context free grammars | Dynamic programming | Hidden markov model | Head drive phrase structure grammar | Support vector machine | Expectation maximization | N best | Maximal likelihood estimation |
| 2004 | Maximum entropy | Expectation maximization | Probabilistic context free grammar | Support vector machine | Decision tree | Hidden markov model | Context free grammars | Dynamic programming | Naive Bayes | Maximal likelihood estimation |
| 2005 | Maximum entropy | Support vector machine | Dynamic programming | Hidden markov model | Naive Bayes | Bootstrapping | N best | Bleu | Expectation maximization | Decision tree |
| 2006 | Support vector machine | Maximum entropy | Dynamic programming | Hidden markov model | Expectation maximization | Bleu | Decision tree | Bootstrapping | N best | Context free grammars |
| 2007 | Support vector machine | Maximum entropy | Bleu | Log linear | Hidden markov model | Expectation maximization | Dynamic programming | Bootstrapping | Perceptron | N best |
| 2008 | Support vector machine | Maximum entropy | Bleu | Hidden markov model | Expectation maximization | Log linear | Dynamic programming | Decision tree | N best | Probabilistic context free grammar |
| 2009 | Support vector machine | Bleu | Maximum entropy | Dynamic programming | Log linear | Expectation maximization | Conditional random fields | Minimum error rate training | Hidden markov model | Bootstrapping |
| 2010 | Support vector machine | Bleu | Expectation maximization | Maximum entropy | Hidden markov model | Dynamic programming | Probabilistic context free grammar | Log linear | Bootstrapping | N best |
| 2011 | Support vector machine | Bleu | Maximum entropy | Expectation maximization | Dynamic programming | Log linear | Cosine similarity | Hidden markov model | Bootstrapping | Minimum error rate training |
| 2012 | Support vector machine | Bleu | Expectation maximization | Conditional random fields | Dynamic programming | Log linear | Maximum entropy | Minimum error rate training | Gibbs sampling | N best |
| 2013 | Support vector machine | Bleu | Graph based | Conditional random fields | Maximum entropy | Cosine similarity | Latent dirichlet allocation | Bag of words | Expectation maximization | Dynamic programming |
| 2014 | Support vector machine | Cosine similarity | Bleu | Expectation maximization | Log linear | Maximum entropy | Perceptron | Conditional random fields | Neural networks | Logistic regression |
| 2015 | Neural networks | Support vector machine | Gradient descent | Stochastic gradient descent | Word2vec | Logistic regression | Bag of words | Cosine similarity | Skip gram | Maximum-entropy |

**Note**: "*" means that there are less than 10 algorithms in this year.